\documentclass{edm_article}

\usepackage{xcolor}
\begin{document}
\title{Student Performance Prediction Using Dynamic Neural Models}
%
\numberofauthors{4}
\author{
{Marina Delianidi}\\
       \affaddr{International Hellenic University, Greece}\\
       \email{dmarina@ihu.gr}
\and{Konstantinos Diamantaras}\\
        \affaddr{International Hellenic University, Greece}\\
        \email{kdiamant@ihu.gr}
\and{George Chrysogonidis}\\
        \affaddr{International Hellenic University, Greece}\\
        \email{g.chrysogonidis@ihu.edu.gr}
\and{Vasileios Nikiforidis}\\
        \affaddr{International Hellenic University, Greece}\\
        \email{v.nikiforidis@ihu.edu.gr} 
}

\maketitle


\begin{abstract}
We address the problem of predicting the correctness of the student's response on the next exam question based on their previous interactions in the course of their learning and evaluation process.
We model the student performance as a dynamic problem and compare the two major classes of dynamic neural architectures for its solution, namely the finite-memory Time Delay Neural Networks (TDNN) and the potentially infinite-memory Recurrent Neural Networks (RNN).
Since the next response is a function of the knowledge state of the student and this, in turn, is a function of their previous responses and the skills associated with the previous questions, we propose a two-part network architecture.
The first part employs a dynamic neural network (either TDNN or RNN) to trace the student knowledge state. The second part applies on top of the dynamic part and it is a multi-layer feed-forward network which completes the classification task of predicting the student response based on our estimate of the student knowledge state.
Both input skills and previous responses are encoded using different embeddings.
Regarding the skill embeddings we tried two different initialization schemes using (a) random vectors and (b) pretrained vectors matching the textual descriptions of the skills.
Our experiments show that the performance of the RNN approach is better compared to the TDNN approach in all datasets that we have used.
Also, we show that our RNN architecture outperforms the state-of-the-art models in four out of five datasets.
It is worth noting that the TDNN approach also outperforms the state of the art models in four out of five datasets, although it is slightly worse than our proposed RNN approach.
Finally, contrary to our expectations, we find that the initialization of skill embeddings using pretrained vectors offers practically no advantage over random initialization. 
\end{abstract}

%


\section{Introduction}
Knowledge is distinguished by the ability to evolve over time. 
This progression of knowledge is usually incremental and its formation is related to the cognitive areas being studied. 
The process of Knowledge Tracing (KT) defined as the task of predicting students' performance has attracted the interest of many researchers in recent decades \cite{corbett1994knowledge}.
The Knowledge State (KS) of a student is the degree of his or her mastering the Knowledge Components (KC) in a certain domain, for example ``Algebra'' or ``Physics''.
A knowledge component generally refers to a learnable entity, such as a concept or a skill, that can be used alone or in combination with other KCs in order to solve an exercise or a problem \cite{koedinger2012knowledge}.
Knowledge Tracing is the process of modeling and assessing a student's KS in order to predict his or her ability to answer the next problem correctly.
The estimation of the student's knowledge state is useful for improving the educational process by identifying the level of his/her understanding of the various knowledge components. By exploiting this information it is possible to suggest appropriate educational material to cover the student's weaknesses and thus maximize the learning outcome.

The main problem of  Knowledge Tracing is the efficient management of the responses over time. 
One of the factors which add complexity to the problem of KT is the student-specific learning pace. 
The knowledge acquisition may differ from person to person and may also be influenced by already existing knowledge. 
More specifically, KT is predominantly considered as a supervised sequence learning problem where the goal is to predict the probability that a student will answer correctly the future exercises, given his or her history of interactions with previous tests.
Thus, the prediction of the correctness of the answer is based on the history of the student's answers in combination with the skill that is currently examined at this time instance.

Mathematically, the KT task is expressed as the probability $P(r_{t+1} = 1|q_{t+1}, X_t)$ that the student will offer the correct response in the next interaction $x_{t + 1}$, where the students learning activities are represented as a sequence of interactions  $X_t = \{ x_1,x_2,x_3,...,x_t\}$ over time $T$. 
The $x_t$ interaction consists of a tuple $(q_t, r_t)$ which represents the question $q_t$ being answered at time $t$ and the student response $r_t$ to the question.
Without loss of generality, we shall assume that knowledge components are represented by skills from a set $S = \{s_1, s_2,..., s_m\}$.
One simplifying assumption, used by many authors \cite{zhang2017dynamic}, is that
every question in the set $Q = \{q_1, q_2,..., q_T\}$ is related to a unique skill from $S$.
Then the knowledge levels of the student for each one of the skills in $S$ compose his or her knowledge state.

The dynamic nature of Knowledge Tracing leads to approa-ches that have the ability to model time-series or sequential data. 
In this work we propose two dynamic machine learning models that are implemented by time-dependent methods, specifically recurrent and time delay neural networks. 
Our models outperform the current state-of-the-art approaches in four out of five benchmark datasets that we have studied. 
The proposed models differ from the existing ones in two main architectural aspects:
\begin{itemize}
    \item we find that attention does not help improve the performance and therefore we make no use of attention layers
    \item we experiment with and compare between two different skill embedding types: (a) initialized by pre-trained embeddings of the textual descriptions of the skill names using standard methods such as Word2Vec and FastText and (b) randomly initialized embeddings based  on skill ids
\end{itemize}


The rest of the paper is organized as follows. 
Section 2 reviews the related works on KT and the existing models for student performance prediction. 
In Section 3 we present our proposed models and describe their architecture and characteristics. 
The datasets we prepared and used are present in Section 4 while the experiments setup and the results are explained in Section 5. 
Finally, Section 6 concludes this work and discusses the future works and extensions of the research.

\section{Related Works}
The problem of knowledge tracing is dynamic as student knowledge is constantly changing over time.
Thus, a variety of methods, highly structured or dynamic, have been proposed to predict students' performance. 
One of the earlier methods is Bayesian Knowledge Tracing (BKT) \cite{corbett1994knowledge} which models the problem as a Hidden Markov chain in order to predict the sequence of outcomes for a given learner.
The Performance Factors Analysis Model (PFA)  \cite{pavlik2009performance} proposed  to tackle the knowledge tracing task by modifying the Learning Factor Analysis model.
It estimates the probability that a student will answer a question correctly by maximizing the likelihood of a logistic regression model. 
The features used in the PFA model, although interpretable, are relatively simple and designed by hand, and may not adequately represent the students' knowledge state \cite{yeung2019deep}.

Deep Knowledge Tracing (DKT) \cite{piech2015deep} is the first dynamic model proposed in the literature utilizing recurrent neural networks (RNN) and specifically the Long Short-Term Memory (LSTM) model \cite{hochreiter1997long} to track student knowledge. 
It uses one-hot encoded skill tags and associated responses as inputs and it trains the neural network to predict the next student response. 
The hidden state of the LSTM can be considered as the latent knowledge state of a student and can carry the information of the past interactions to the output layer. 
The output layer of the model computes the probability of the student answering correctly a question relating to a specific Knowledge Component.

Another approach for predicting student performance is the Dynamic Key-Value Memory Network (DKVMN) \cite{zhang2017dynamic} which relies on an extension of \textit{memory networks} proposed in \cite{miller2016key}.
The model tries to capture the relationship between different concepts.
The DKVMN model outperforms DKT using memory slots as key and value components to encode the knowledge state of students. 
Learning or forgetting of a particular skill are stored in those components and controlled by read and write operations through the Least Recently Used Access (LRUA) attention mechanism \cite{santoro2016meta}.
The key component is responsible for storing the concepts and is fixed during testing while the value component is updated when a concept state changes. 
The latter means that when a student acquires a concept in a test the value component is updated based on the correlation between exercises and the corresponding concept.

The Deep-IRT model \cite{yeung2019deep} is the newest approach that extends the DKVMN model. 
The author combined the capabilities of DKVMN with the Item Response Theory (IRT) \cite{hambleton1991fundamentals} in order to measure both student ability and question difficulty. 
At the same time, another model, named Sequential Key-Value Memory Networks (SKVMN) \cite{abdelrahman2019knowledge}, tried to overcome the problem of DKVMN to capture long term dependencies in the sequences of exercises and generally in sequential data. 
This model combines the DKVMN mechanism with the Hop-LSTM, a variation of LSTM architecture and has the ability to discover sequential dependencies among exercises, but it skips some LSTM cells to approach previous concepts that are considered relevant.
Finally, another newly proposed model is Self Attentive Knowledge Tracing (SAKT) \cite{pandey2019self}. 
SAKT utilizes a self-attention mechanism and mainly consists of three layers: an embedding layer for interactions and questions followed by a Multi-Head Attention layer \cite{vaswani2017attention} and a feed-forward layer for student response prediction. 

The above models either use simple features (e.g. PFA) or they use machine learning approaches such as key-value memory networks or attention mechanisms that may add significant complexity. 
However we will show that similar and often, in fact, better performance can be achieved by simpler dynamic models combining embeddings and recurrent and/or time-delay feed-forward networks as proposed next.

\section{Proposed Approach}

\subsection{Dynamic Models}
As referenced in the relative literature, knowledge change over time is often modeled by dynamic neural networks.
The dynamic models produce output based on a time window, called ``context window'', that contains the recent history of inputs and/or outputs.

\begin{figure}[h!]
  \centering
  \includegraphics[scale=0.6]{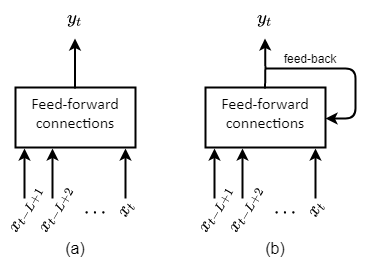}
  \caption{Dynamic model architectures: (a) Time-Delay Neural Network (b) Recurrent Neural Network.}
  \label{fig:DNN_architecture}
\end{figure}

There are two types of dynamic neural networks (Figure \ref{fig:DNN_architecture}):
(a) Time-Delay Neural Networks (TDNN), with only feed-forward connections and finite-memory of length $L$ equal to the length of the context window, and
(b) Recurrent Neural Networks (RNN) with feed-back connections that can have potentially infinite-memory although, practically, their memory length is dictated by a forgetting factor parameter.

\subsection{The Proposed Models}

We approach the task of predicting the student response (0=wrong, 1=correct) on a question involving a specific skill as a dynamic binary classification problem.
In general, we view the response $r_t$ as a function of the previous student interactions:
\begin{equation}
    r_t = h( q_t,q_{t-1},q_{t-2},\dots,r_{t-1},r_{r-2},\dots ) + \epsilon_t
    \label{eq:total_recurrent_model}
\end{equation}
where $q_t$, is the skill tested on time $t$ and $\epsilon_t$ is the prediction error. The response is therefore a function of the current and the previous tested skills $\{q_t, q_{t-1}, q_{t-2}, \dots\}$, as well as the previous responses $\{r_{t-1}, r_{t-2}, \dots\}$ given by the student.

\begin{figure}[h!]
  \centering
  \includegraphics[scale=0.62]{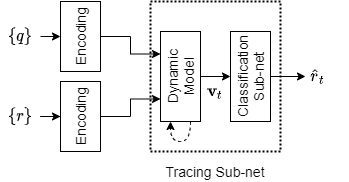}
  \caption{General proposed architecture. The dynamic model can be either a Recurrent Neural Network (with a feedback connection from the output of the dynamic part into the model input) or a Time Delay Neural Network (without feedback connection).}
  \label{fig:EDM_architecture}
\end{figure}

We implement $h$ as a dynamic neural model.
Our proposed general architecture is shown in Figure \ref{fig:EDM_architecture}.
The inputs are the skill and response sequences $\{q\}$, $\{r\}$ collected during a time-window of length $L$ prior to time $t$.
Note that the skill sequence includes the current skill $q_t$ but the response sequence does not contain the current response which is actually what we want to predict.
The architecture consists of two main parts:
\begin{itemize}
    \item The Encoding sub-network. It is used to represent the response and skill input data using different embeddings.
    Clearly, embeddings are useful for encoding skills since skill ids are categorical variables. 
    We found that using embeddings to encode responses is also very beneficial.
    The details of the embeddings initialization and usage are described in the next section.
    \item The Tracing sub-network. This firstly estimates the knowledge state of the student and then uses it to predict his/her response.
    Our model function consists of two parts: (i) the Knowledge-Tracing part, represented by the dynamic model $f$, which predicts the student knowledge state $\mathbf{v}_t$ and (ii) the classification part $g$, which predicts the student response based on the estimated knowledge state:
    \begin{eqnarray}
        \mathbf{v}_t &=& f(q_t,q_{t-1},q_{t-2},\dots,r_{t-1},r_{r-2},\dots)
        \label{eq:kt_estimation}
        \\
        \hat{r}_t &=& g(\mathbf{v}_t)
        \label{eq:classification}
    \end{eqnarray}
    Depending on the memory length, we obtain two categories of models:
    \begin{itemize}
        \item[(a)] models based on RNN networks which can potentially have infinite memory.
        In this case the KT model is recurrent:
        \[
        \mathbf{v}_t = f(\mathbf{v}_{t-1}, q_t,q_{t-1},\dots,q_{t-L},r_{t-1},\dots,r_{r-L})
        \]
        \item[(b)] models based on TDNN networks which have finite memory of length $L$.
        In this case the KT model has finite impulse response $L$:
        \[
        \mathbf{v}_t = f(q_t,q_{t-1},\dots,q_{t-L},r_{t-1},\dots,r_{r-L})
        \]
    \end{itemize}
\end{itemize} 
Although RNNs have been used in the relevant literature, it is noteworthy that TDNN approaches have not been investigated in the context of knowledge tracing. 
The classification part is modeled by a fully-connected feed-forward network with a single output unit.

We investigated two different architectures: one based on recurrent neural networks and another based on time delay neural networks.
The details of each proposed model architecture are described below.

\subsection{Encoding Sub-network}
The first part in all our proposed models consists of two parallel embedding layers with dimensions $d_q$ and $d_r$, respectively, which encode the tested skills and the responses given by the student.
During model training the weights of the Embedding layers are updated. 
The response embedding vectors are initialized randomly.
The skill embedding vectors, on the other hand, are initialized either randomly or using pretrained data. In the latter case we use pretrained vectors corresponding to the skill names obtained from Word2Vec \cite{mikolov2013efficient} or FastText \cite{joulin2016fasttext} methods.

A 1D spatial dropout layer \cite{tompson2015efficient} is added after each Embedding layer. 
The intuition behind the addition of spatial dropout was the overfitting phenomenon that was observed in the first epochs of each validation set. We postulated that the correlation among skill name embeddings, that might not actually exist, confused the model.

\subsection{Tracing Sub-network}

We experimented with two types of main dynamic sub-net- 
works, namely Recurrent Neural Networks and Time Delay Neural Networks. These two approaches are described next.

\subsubsection{RNN Approach: Bi-GRU Model}

\begin{figure}[h!]
  \centering
  \includegraphics[scale=0.525]{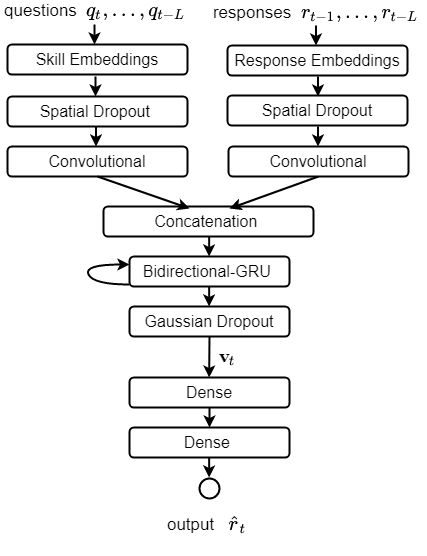}
  \caption{Bi-GRU model}
  \label{fig:Bi_GRU}
\end{figure}

The model architecture based on the RNN method for the knowledge tracing task is shown in Figure \ref{fig:Bi_GRU}.

The Spatial Dropout rate following the input embedding layers is $0.2$ for most of used datasets.
Next, we feed the skills and the responses input branches into a Convolutional layer consisting of 100 filters, with kernel size 3, stride 1, and ReLU activation function.
The Convolutional layer acts as a projection mechanism that reduces the input dimensions from the previous Embedding layer.
This is found to help alleviate the overfitting problem.
To the best of our knowledge, Convolutional layers have not been used in previously proposed neural models for this task.
The two input branches are then concatenated to feed a Bidirectional Gated Recurrent Unit (GRU) layer with 64 units \cite{cho2014learning}.
Batch normalization and ReLU activation layers are applied between convolutional and concatenation layers.
This structure has resulted after extensive experiments with other popular recurrent models such as LSTM, plain GRU and also bi-directional versions of those models and we found this to be the proposed architecture is the most efficient one.
On top of the RNN layer we append a fully connected sub-network  consisting of three dense layers with 50 and 25 units and one output unit respectively.
The first two dense layers have a ReLU activation function while the last one has sigmoid activation which is used to make the final prediction ($0 < \hat{r}_t < 1$).

\subsubsection{TDNN Approach}

\begin{figure}[h!]
  \centering
  \includegraphics[scale=0.525]{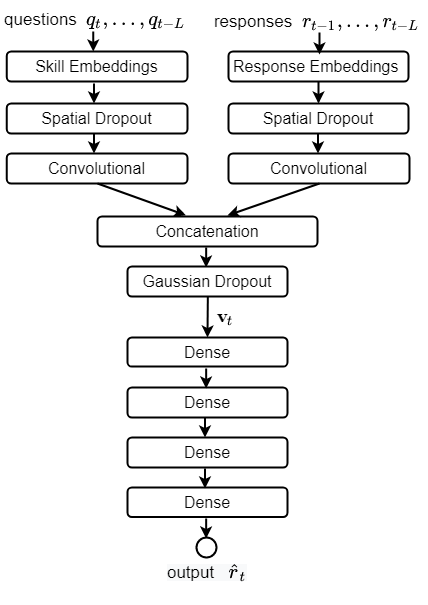}
  \caption{TDNN model}
  \label{fig:Tdnn_v1}
\end{figure}

In our TDNN model (Figure \ref{fig:Tdnn_v1}) we add a Convolutional layer after each embedding layer with 50 filters and kernel size equal to 5. 

Batch normalization is used before the ReLU activation is applied.
As with the RNN model, the two input branches are concatenated to feed the classification sub-network.
It consists of four dense layers with 20, 15, 10, and 5 units respectively, using the ReLU activation function.
This funnel schema of hidden layers (starting with wider layers and continuing with narrower ones) has helped achieve better results for all datasets we have experimented with. 
In the beginning of the classification sub-network we insert a Gaussian Dropout layer \cite{srivastava2014dropout} which multiplies neuron activations with a Gaussian random variable of mean value 1. This has been shown to work as good as the classical Bernoulli noise dropout and in our case even better.

\section{Datasets}

\begin{table*} [htb]
\centering
  \caption{Datasets Overview.}
  \label{tab:datasets}
    \begin{tabular}{|c|c|c|c|c|}
    \hline
    Dataset & Skills &  Students & Responses & Baseline Accuracy \\ \hline
    ASSISTment09 & 110 & 4,151 & 325,637 & 65.84\%\\ \hline
    ASSISTment09 corrected & 101 & 4,151 & 274,590 & 66.31\%\\ \hline
    ASSISTment12 & 196 & 28,834 & 2,036,080 & 69.65\%\\ \hline
    ASSISTment17 & 101 & 1,709 & 864,713 & 62.67\%\\ \hline
    FSAI-F1toF3 & 99 & 310 & 51,283 & 52.98\%\\
    \hline
  \end{tabular}
\end{table*}

We tested our models using four popular datasets from the ASSISTments online tutoring platform.
Three of them, ``\textit{ASSISTment09}'',  ``\textit{ASSISTment09 corrected}''\footnote{\rule{0pt}{1.2\baselineskip}https://sites.google.com/site/assistmentsdata/home/assist
ment-2009-2010-data/skill-builder-data-2009-2010},
and ``\textit{ASSISTment12}''\footnote{https://sites.google.com/site/assistmentsdata/home/2012-13-school-data-with-affect} were provided by the above platform.
The fourth dataset, named ``\textit{ASSISTment17}'' was obtained from 2017 Data Mining competition page\footnote{https://sites.google.com/view/assistmentsdatamining/dat a-mining-competition-2017}.
Finally a fifth dataset, ``\textit{FSAI-F1toF3}'' provided by ``Find Solution Ai Limited'' was also used in our experiments.
It is collected using data from the from the 4LittleTrees\footnote{https://www.4littletrees.com} adaptive learning application.

\subsection{Datasets Descriptions}

The ASSISTments datasets contain data from student tests on mathematical problems 
\cite{assistmentsdata} and the content is organized in columns style.
The student's interaction is recorded on each line.
There are one or more interactions recorded for each student. 
We take into account the information concerning the responses of students to questions related with a skill.
Thus, we use the following columns:
``\textit{user\_id}'', ``\textit{skill\_id}'',
``\textit{skill\_name}'', 
and ``\textit{correct}''. 
The ``\textit{skill\_name}'' contains a verbal description of the skill tested.
The ``\textit{correct}'' column contains the values of the students' responses which are either $1$ (for correct) or $0$ (for wrong).

The original ``\textit{ASSISTment09}'' dataset contains 525,534 student responses.
It has been used extensively in the KT task from several researchers but according to \cite{assistmentsdata} data quality issues have been detected concerning duplicate rows. 
In our work we used the ``\textit{preprocessed ASSISTment09}'' dataset found on DKVMN\footnote{https://github.com/jennyzhang0215/DKVMN} and Deep-IRT\footnote{https://github.com/ckyeungac/DeepIRT} models GitHubs. 
In this dataset  the duplicate rows and the empty field values were cleaned, so that finally 1,451 unique students participate with 325,623 total responses and 110 unique skills.

Even after this cleaning there are still some problems such as duplicate skill ids for the same skill name.
These problems have been corrected in the ''\textit{Assistment09 corrected}'' dataset.
This dataset contains 346,860 students interactions and has been recently used in \cite{xu2020dynamic}.

The ``\textit{ASSISTment12}'' dataset contains students' data until the school year 2012-2013. The initial dataset contains 6,123,270 responses and 198 skills. 
Some of the skills have the same skill name but different skill id.
The total number of skill ids is 265. 
The ``\textit{Assistment17}'' dataset contains 942,816 students responses and 101 skills. 

Finally, the ``\textit{FSAI-F1toF3}'' dataset is the smallest dataset we used. 
It involves responses to mathematical problems from 7th grade to 9th grade Hong Kong students and consists of 51,283 students responses from 310 students on 99 skills and 2,266 questions. As it is commonly the case in most studies using this dataset, we have used the question tag as the model input $q_t$.

\subsection{Data Preprocessing}

No preprocessing was performed on the ``\textit{ASSISTment09}'' and ``\textit{FSAI-F1toF3}'' datasets.
For the remaining datasets we followed three preparation steps. 

First, the skill ids had been repaired by replacement. 
In particular, the ``\textit{ASSISTments09 corrected}'' dataset contained skills of the form of ``\textit{skill1\_skill2}'' and ``\textit{skill1\_skill2\_skill3}'' which correspond to the same skill names, so we have merged them into the first skill id, found before the underscore. 
In other words, the skill ``\textit{10\_13}'' was replaced with skill ``\textit{10}'' and so on.
Moreover, few misspellings were observed that were corrected and the punctuations found in three skill names were converted to the corresponding words. 
For example, in the skill name ``\textit{Parts of a Polnomial Terms Coefficient Monomial Exponent Variable}'' we corrected the ``\textit{Polnomial}'' with ``\textit{Polynomial}''. Also, in the skill name ``\textit{Order of Operations +,-,/,*() positive reals}'' we replaced the symbols ``\textit{+,-,/,* ()}'' with the words that express these symbols, ie. ``\textit{addition subtraction division multiplication parentheses}''. 
The latter preprocessing action was preferred over the removal of punctuations since the datasets referred to mathematical methods and operations and without them, we would lose the meaning of each skill.  
Similar procedure has been followed for the ``\textit{ASSISTments12}'' dataset. 
Furthermore, spaces after some skill names were removed  i.e. the skill name ``\textit{Pattern Finding  }'' became ``\textit{Pattern Finding}''. 
In the ``\textit{ASSISTment17}'' dataset we came across skill names as ``\textit{application: multi-column subtraction}'' and corrected them by replacing punctuation marks such as ``\textit{application multi column subtraction}''.
That text preparation operations made to ease the generation of word embeddings of the skill names descriptions.
In addition, in the ``\textit{ASSISTment17}'' dataset, the problem ids are used instead of the skill ids. 
We had to match and replace the problem ids with the corresponding skill ids with the aim of uniformity of the datasets between them.

Secondly, all rows containing missing values were discarded. 
Thus, after the preprocessing, the statistics of the data sets were formulated as described in the Table \ref{tab:datasets}.

Finally, we split the datasets so that 70\% was used for training and 30\% for testing.
Then, the training subset was further split into five train-validation subsets using 80\% for training and 20\% for validation. 

\section{Experiments}

In this section we experimentally validate the effectiveness of the proposed methods by comparing them with each other and also with other state-of-the-art performance prediction models.
The Area Under the ROC Curve (AUC) \cite{ling2003auc} metric is used for comparing the predicting probability correctness of student's response. 

The state-of-the-art knowledge tracing models we are compared with the DKT, DKVMN and Deep-IRT. 
We performed the experiments for our proposed models Bi-GRU, TDNN as well as for each of the previous model for all datasets, using the code provided by the authors on their GitHubs. 
It is worth noting that the python GitHub code\footnote{\rule{0pt}{1.2\baselineskip}https://github.com/lccasagrande/Deep-Knowledge-Tracing} used for the DKT model experiments requires the entire dataset file and the train/test splitting is performed during the code execution. 

All the experiments were performed on a workstation with Ubuntu operating system, Intel i5 CPU and 16GB Titan Xp GPU card.

\begin{table*}[htb]
    \caption{Models experiments settings} 
    \label{tab:settings}
    \centering
    \begin{tabular}{|l | r|  r| }
        \hline  
            \multicolumn{1}{|c|}{Parameters} & 
            \multicolumn{1}{c|}{Bi-GRU} & 
            \multicolumn{1}{c|}{TDNN}\\  \hline
            Learning rate & 0.001 & 0.001 \\ \hline
            Learning rate schedule & yes & no\\ \hline
            Training epochs& 30 & 30 \\ \hline
            Batch size & 32 & 50 \\ \hline
            Optimizer & Adam & AdaMax\\ \hline
            History window length & 50 & 50 \\\hline
            Skill embeddings dim. & 100 \& 300 & 100 \& 300\\ \hline
            Skill embeddings type & Random, W2V, FastText & Random, W2V, FastText\\ \hline
            Responses embeddings dim. & Same to skill dim. & Same to skill dim.\\ \hline
            Responses embeddings type & Random & Random\\
        \hline   
\end{tabular}
\end{table*}
\subsection{Skill embeddings initialization}

As mentioned earlier, skill embeddings are initialized either randomly or using pretrained vectors.
Regarding the initialization of the skill embeddings with pretrained vectors we used two methods described next. 
In first method we used the text files from Wikipedia2Vec\footnote{https://wikipedia2vec.github.io/wikipedia2vec/} \cite{yamada2020wikipedia2vec} that is based on Word2Vec method and contains pretrainable embeddings for the  word representation vectors in English language in 100 and 300 dimensions.
In second method we used the ``\textit{SISTER}'' (SImple SenTence EmbeddeR)\footnote{https://pypi.org/project/sister/} library to prepare the skill name embeddings  based on FastText in 300 dimensions pretrained word embeddings.
Each skill name consists of one or more words. 
Thus, for the Word2Vec method, the skill name embeddings vector is created by adding the word embeddings vectors, while in case of FastText, the skill name embeddings are created by taking the average of the word embeddings. 

Especially for the FsaiF1toF3 dataset, the question embeddings are initialized either randomly or using the pretrained word representations of the corresponding skill descriptions by employing the Wikipedia2Vec and SISTER methods as described above. Since many questions belong to the same skill, in this case the corresponding rows in the embedding matrix are initialized by the same vector.

\subsection{Experimental Settings}
We performed the cross-validation method for the 5 training and validation set pairs. 
This was to choose the best architecture and parameter settings for each of the proposed models.
Using the train and test sets we evaluated the chosen architectures for all the datasets. 

One of the basic hyperparameters of our models that affect to the inputs is the $L$.  It represents the student's interaction history window length. 
The inputs with $L$ sequence of questions and $L-1$ sequence of responses. 
The best results we succeeded are when using $L=50$ for the both Bi-GRU and TDNN models.
The batch sizes used in the models during the training are: 32 in Bi-GRU and 50 in TDNN. 

Since specific dimensions of the pretrained word embeddings are provided, we used the same dimensions in case of random embedding in order to take the comparable results. 
Skill embeddings and responses embeddings set in the same dimensions.   

The scheduler learning rate is implementing in Bi-GRU starting from 0.001 
and reducing over the training operation of the models that performs for 30 epochs.
During training we applied the following learning rate schedule depending on the epoch number $n$:
\[
    lr = \left\{
        \begin{array}{ll}
        r_{init} &
        \text{if } n < 10
        \\
        r_{init} \times e^{(0.1\cdot(10-n))} &
        \text{otherwise}
        \end{array}
    \right.
\]

In case of the TDNN-based model, the learning rate equals 0.001 and is the same during the whole training process for 30 epochs.
We used cross-entropy optimization criterion and the Adam or AdaMax \cite{kingma2014adam} learning algorithms.
 
Dropout with rate = $0.2$ or $0.9$ is also applied to the Bi-GRU model while the dropout rate of the TDNN equals to one of the  $(0.2, 0.4, 0.6, 0.9)$ values through to the Gaussian dropout layer.
We observed a reduction of overfitting during model training by changing the Gaussian dropout rate relative to the dataset's size.
Thus, the smaller dataset size is, the bigger dropout rate has been used.

The various combinations of parameters settings were applied during the experimental process for all proposed models presented in Table \ref{tab:settings}.

\subsection{Experimental Results}

\begin{table*}[htb]
    \centering
    \caption{Comparison between our proposed models - AUC (\%). (R) = random skill embedding initialization, (W) = skill embedding initialization using W2V, (F) = skill embedding initialization using FastText. Datasets: (a) \textit{ASSISTment09}, (b) \textit{ASSISTment09 corrected}, (c) \textit{ASSISTment12}, (d) \textit{ASSISTment17}, (e) \textit{FSAI-F1toF3}}
    \label{tab:compare_our_models} 
    \begin{tabular}{|c|c|c|c|c|c|}
    \hline
         & $d_q=100$(R) & $d_q=300$(R) & $d_q=100$(W) & $d_q=300$(W) & $d_q=300$(F)
         \\
        \hline
        Bi-GRU & 82.55 & 82.45 & 82.52 & 82.55 & 82.39
        \\ \hline
        TDNN & 81.54 & 81.67 & 81.59 & 81.50 & 81.53
        \\
        \hline
    \end{tabular}
    \\
    (a)
    \\
    ~\\
    \begin{tabular}{|c|c|c|c|c|c|}
        \hline
         & $d_q=100$(R) & $d_q=300$(R) & $d_q=100$(W) & $d_q=300$(W) & $d_q=300$(F)
         \\
        \hline
        Bi-GRU & 75.27 & 75.13 & 75.14 & 75.09 & 75.12
        \\ \hline
        TDNN & 74.38 & 74.39 & 74.40 & 74.33 & 74.37
        \\
        \hline
    \end{tabular}
    \\
    (b)
    \\
    ~\\
        \begin{tabular}{|c|c|c|c|c|c|}
        \hline
         & $d_q=100$(R) & $d_q=300$(R) & $d_q=100$(W) & $d_q=300$(W) & $d_q=300$(F)
         \\
        \hline
        Bi-GRU & 68.37 & 68.37 & 68.40 & 68.23 & 68.27
        \\ \hline
        TDNN & 67.95 & 67.97 & 67.99 & 67.95 & 67.91
        \\
        \hline
    \end{tabular}
    \\
    (c)
    \\
    ~\\
        \begin{tabular}{|c|c|c|c|c|c|}
        \hline
         & $d_q=100$(R) & $d_q=300$(R) & $d_q=100$(W) & $d_q=300$(W) & $d_q=300$(F)
         \\
        \hline
        Bi-GRU & 73.62 & 73.58 & 73.76 & 73.54 & 73.58
        \\ \hline
        TDNN & 71.68 & 71.75 & 71.52 & 71.81 & 71.83
        \\
        \hline
    \end{tabular}
    \\
    (d)
    \\
    ~\\
        \begin{tabular}{|c|c|c|c|c|c|}
        \hline
         & $d_q=100$(R) & $d_q=300$(R) & $d_q=100$(W) & $d_q=300$(W) & $d_q=300$(F)
         \\
        \hline
        Bi-GRU & 70.47 & 69.34 & 70.24 & 69.80 & 69.51
        \\ \hline
        TDNN & 70.03 & 69.80 & 69.80 & 70.11 & 70.06 
        \\
        \hline
    \end{tabular}
    \\
    (e)
\end{table*}

\begin{table*}[htb]
    \caption{Comparison test results of evaluation measures - the AUC metric (\%) }
    \label{tab:results}
    \centering
    \begin{tabular}{|l|c|c|c|c|c|}
        \hline
            Dataset & DKT & DKVMN & Deep-IRT & Bi-GRU & TDNN\\ \hline
    ASSISTment09 & 81.56\% & 81.61\% & 81.65\% &\textbf{82.55}\%$^{(1,2)}$ & 81.67\%$^{(3)}$ \\ \hline
    ASSISTment09 corrected & 74.27\% & 74.06\% & 73.41\% & \textbf{75.27}\%$^{(1)}$ & 74.40\%$^{(2)}$\\ \hline
    ASSISTment12 & 69.40\% & 69.26\% & \textbf{69.73}\% & 68.40\%$^{(4)}$ & 67.99\%$^{(4)}$ \\ \hline
    ASSISTment17 & 66.85\% & 70.25\% & 70.54\% & \textbf{73.76}\%$^{(4)}$ & 71.83\%$^{(5)}$ \\ \hline
    FSAI-F1toF3 & 69.42\% & 68.40\% & 68.69\% & \textbf{70.47}\%$^{(1)}$ & 70.11\%$^{(2)}$ \\
    \hline
    \multicolumn{6}{l}{
    \rule{0pt}{3ex}
    $^{(1)}$ $d_q=d_r=100$, Random, ~~
    $^{(2)}$ $d_q=d_r=300$, W2V, ~~
    $^{(3)}$ $d_q=d_r=300$, Random,
    }
    \\
    \multicolumn{6}{l}{
    $^{(4)}$ $d_q=d_r=100$, W2V, ~~
    $^{(5)}$ $d_q=d_r=300$, FastText
    }
    \end{tabular}
\end{table*}

The experiments results of our models are shown in Table \ref{tab:compare_our_models}. Comparing our models with each other we can see that the RNN-based Bi-GRU model outperforms the TDNN-based model in all datasets. 
It achieved best results when 100d embeddings were used either in pretrained or the random initialization type. 

We observed that in both Bi-GRU or TDNN, the embedding type is not the significant parameter that affects the models performance. 
The differences between the results of the experiments showed that the size of embeddings dimensions not particularly contributed to the final result and the difference in performance of the models was small.

Except for our models, we performed experiments for all datasets on the previous models we compared. 
For three of the datasets, specifically  for ``\textit{ASSISTment09 corrected}'', ``\textit{ASSISTment12}'' and ``\textit{ASSISTment17}'' there were not available results in the corresponding papers. 
In this paper, we present the results of the experiments we run using that models codes.

The best experimental results of the ours models in comparison with the previous models for each dataset are presented in Table \ref{tab:results}.
The model that has the best performance for the four of datasets is the Bi-GRU. 
Except for that, the TDNN-based model has better performance in comparison to the previous models for four datasets.
The only dataset, for which the previous models overcomed our models is the ``\textit{ASSISTment12}''.

\subsection{Discussion}

Our model architecture is loosely based on the DKT model and offers improvements in the aspects discussed below.
First, we employ embeddings for representing both skills and responses.
It is known that embeddings offer more useful representations compared to one-hot encoding because they can capture the similarity between the items they represent \cite{wang2020survey}.
Second, we thoroughly examined dynamical neural models for estimating the student knowledge state by trying both infinite-memory RNNs and finite-memory TDNNs.
To our knowledge, TDNNs have not been well studied in the literature with respect to this problem.
Third, we used convolutional layers in the inputs encoding sub-net. We found that this layer functioned as a reducing mechanism of the embedding dimensions and in conjunction with the dropout layer mitigated the overfitting problem.  
The use of Convolutional layers is a novelty in models tackling the knowledge tracing problem. 
Fourth, unlike DKT, we used more hidden layers in the classification sub-net.
Our experiments demonstrate that this gives more discriminating capability to the classifier and improves the results.
Finally, our experiments with key-value modules and attention mechanism did not help further improve our results and so these experiments are not reported here.
In the majority of the datasets we examined our model outperforms the state-off the models employing key-value mechanisms such as DKVMN and Deep-IRT.

\begin{table*}[hbt]
\centering
    \caption{Statistical significance testing results of Bi-GRU and TDNN}
    \label{tab:p_value}
    \begin{tabular}{|l|c|c|}
    \hline
    Dataset  & P-value \\ \hline
    ASSISTment09 & 7.34 e-59 \\ \hline
    ASSISTment09 corrected & 2.31 e-52 \\ \hline
    ASSISTment12 & 1.45 e-203 \\ \hline
    ASSISTment17 & 7.96 e-44   \\ \hline
    FSAI-F1toF3  & 1.38 e-84 \\ \hline
 \end{tabular}
\end{table*}

In addition to the AUC metric which is typically used for evaluating the performance of our machine learning models, we applied statistical significance testing to check the similarity between out Bi-GRU and TDNN models.
Specifically, we performed a T-Test between the outcomes of the two models in all training data using the best configuration settings as shown in Table \ref{tab:results}.
The results reported in Table \ref{tab:p_value} show that the P-value calculated in all cases is practically zero which proves the hypothesis that the two models are significantly different.

\section{Conclusion and Future Work}

In this paper we propose a novel two-part neural network architecture for predicting student performance in the next exam or exercise based on their performance in previous exercises.
The first part of the model is a dynamic network which tracks the student knowledge state and the second part is a multi-layer neural network classifier.
For the dynamic part we tested two different models: a potentially infinite memory recurrent Bidirectional GRU model and a finite memory Time-Delay neural network (TDNN). 
The experimental process showed that the Bi-GRU model achieves better performance compared to the TDNN model.
Despite the fact that TDNN models have not been used for this problem in the past, our results have shown that they can be just as efficient or even better compared to previous state-of-art RNN models and only slightly worse than our proposed RNN model.
The model inputs are the student's skills and responses history which are encoded using embedding vectors. Skill embeddings are initialized either randomly or by pretrained vectors representing the textual descriptions of the skills.
A novel feature of our architecture is the addition of spatial dropout and convolutional layers immediately after the embeddings layers.
These additions have been shown to reduce the overfitting problem.
We found that the choice of initialization of the skill embeddings has little effect on the outcome of our experiments.
Moreover, noting that there is a different use of the same datasets in different studies, we described in detail the process of the datasets pre-processing, and we provide the  train, validation and test splits of the data that were used in our experiments on our GitHub repository\footnote{\rule{0pt}{1.2\baselineskip} https://github.com/delmarin35/Dynamic-Neural-Models-for-Knowledge-Tracing}.
The extensive experimentation with more benchmark datasets as well as the study of variants of the proposed models will be the subject of our future work with the aim of even further improving the prediction performance of the models.


\section{Acknowledgments}

We would like to thank NVIDIA Corporation for the kind donation of an Titan Xp GPU card that was used to run our experiments.


%
\bibliographystyle{abbrv}
\bibliography{edm_refs}  
%
%

\balancecolumns
\end{document}